\documentclass{article}

\usepackage{microtype}
\usepackage{graphicx}
\usepackage{subfigure}
\usepackage{booktabs} 
\usepackage{lipsum}
\usepackage{caption}

\usepackage{hyperref}

\newcommand{\D}{\mathbf{D}}
\newcommand{\U}{\mathbf{U}}

\usepackage[accepted]{icml2024}

\usepackage{amsmath}
\usepackage{amssymb}
\usepackage{mathtools}
\usepackage{amsthm}
\usepackage{kotex}
\usepackage{xcolor}
\usepackage{listings}
\usepackage{color}
\usepackage{color}
\usepackage{graphicx}
\definecolor{deepblue}{rgb}{0,0,0.5}
\definecolor{deepred}{rgb}{0.6,0,0}
\definecolor{deepgreen}{rgb}{0,0.5,0}
\DeclareFixedFont{\ttb}{T1}{txtt}{bx}{n}{8} 
\DeclareFixedFont{\ttm}{T1}{txtt}{m}{n}{8}  
\usepackage{listings}

\newcommand\pythonstyle{\lstset{
language=Python,
basicstyle=\ttm,
morekeywords={self},              
keywordstyle=\ttb\color{deepblue},
emph={MyClass,__init__},          
emphstyle=\ttb\color{deepred},    
stringstyle=\color{deepgreen},
frame=tb,                         
showstringspaces=false
}}

\usepackage[capitalize,noabbrev]{cleveref}

\theoremstyle{plain}

\theoremstyle{definition}

\theoremstyle{remark}

\lstnewenvironment{python}[1][]
{
\pythonstyle
\lstset{#1}
}
{}

\usepackage[textsize=tiny]{todonotes}

\begin{document}

\twocolumn[
\icmltitle{Upsample Guidance: Scale Up Diffusion Models without Training}



\icmlsetsymbol{equal}{*}

\begin{icmlauthorlist}
\icmlauthor{Juno Hwang}{snu}
\icmlauthor{Yong-Hyun Park}{snu}
\icmlauthor{Junghyo Jo}{snu,snuai,kias}
\end{icmlauthorlist}

\icmlaffiliation{snu}{Department of Physics Education, Seoul National University, Seoul, Korea}
\icmlaffiliation{snuai}{Center for Theoretical Physics and Artificial Intelligence Institute, Seoul National University, Seoul, Korea}
\icmlaffiliation{kias}{School of Computational Sciences, Korea Institute for Advanced Study, Seoul, Korea}

\icmlcorrespondingauthor{Junghyo Jo}{jojunghyo@snu.ac.kr}

\icmlkeywords{Machine Learning, ICML}

\vskip 0.3in
]



\printAffiliationsAndNoticeForArxiv{} 

\begin{abstract}
Diffusion models have demonstrated superior performance across various generative tasks including images, videos, and audio. 
However, they encounter difficulties in directly generating high-resolution samples. 
Previously proposed solutions to this issue involve modifying the architecture, further training, or partitioning the sampling process into multiple stages. 
These methods have the limitation of not being able to directly utilize pre-trained models as-is, requiring additional work.  
In this paper, we introduce \textit{upsample guidance}, a technique that adapts pretrained diffusion model (e.g., $512^2$) to generate higher-resolution images (e.g., $1536^2$) by adding only a single term in the sampling process.
Remarkably, this technique does not necessitate any additional training or relying on external models. 
We demonstrate that upsample guidance can be applied to various models, such as pixel-space, latent space, and video diffusion models.
We also observed that the proper selection of guidance scale can improve image quality, fidelity, and prompt alignment.
\end{abstract}

\begin{figure}
\begin{center}
    \centerline{\includegraphics[width=0.8\columnwidth ]{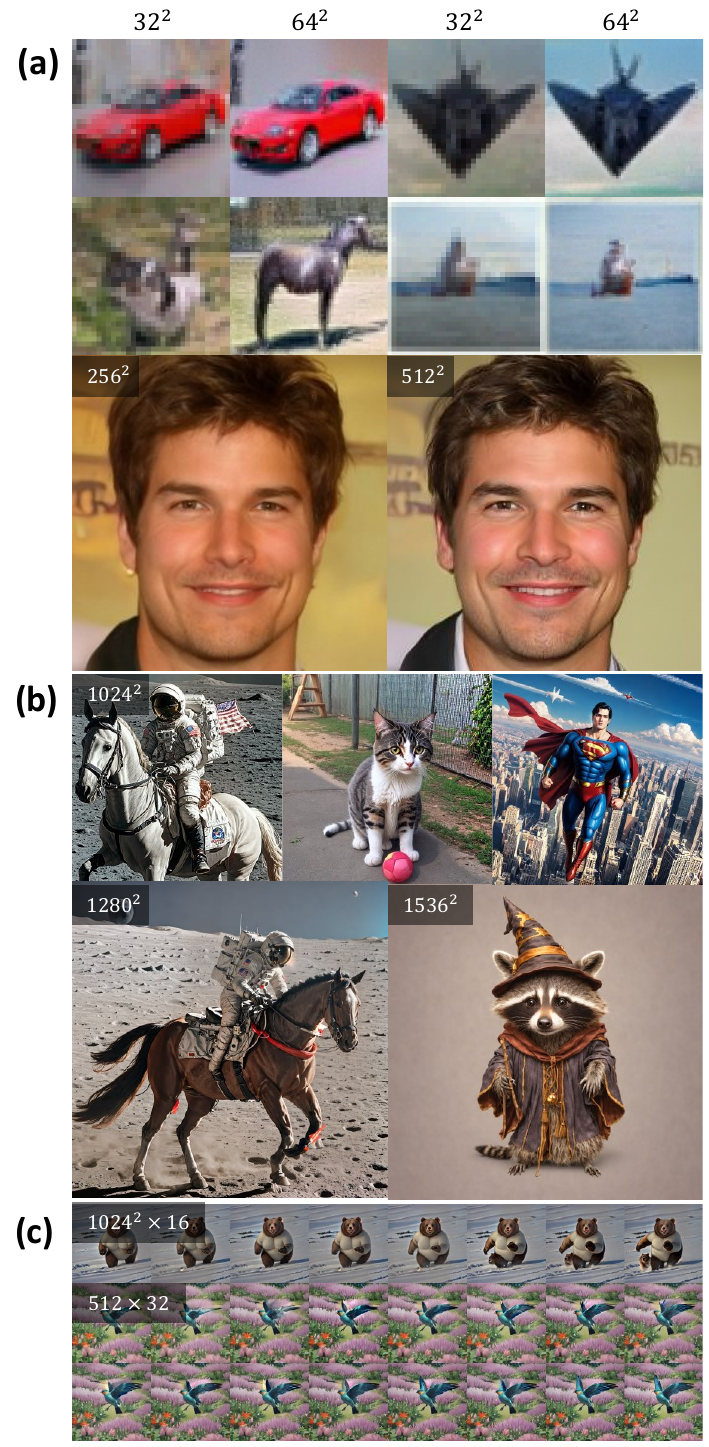}}
    \caption{
    High-resolution samples with upsample guidance. The original trained resolution is increased ($\geq 2$ times) through upsample guidance. (a) Images sampled at twice the resolution for the models trained on CIFAR-10 and CelebA-HQ datasets at $32^2$ and $256^2$ resolutions, respectively. The adjacent image pairs are sampled from the same initial noise. (b) High-resolution images of latent diffusion models using upsample guidance.
    (c) Upsampled snapshots of text-to-video models. The upper panel represents spatial upsampling, while the lower panel represents temporal upsampling.
    }
    \label{fig:showcase}
\end{center}
\vskip -0.2in    
\end{figure}

\section{Introduction}

Diffusion models are generative models that generate samples by progressively restoring the original distribution from prior noise distribution, by modeling the reverse process of data diffusion. \citep{ho2020denoising, song2020denoising, song2020score} Recently, diffusion models have demonstrated state-of-the-art performance in various domains such as image \citep{ho2022imagen, ramesh2022hierarchical, rombach2022high, podell2023sdxl}, video \citep{ho2022video, guo2023animatediff, blattmann2023videoldm, blattmann2023stable}, audio, and 3D generation \citep{poole2022dreamfusion}.

Despite its effectiveness in image generation, generating high-resolution images remains a challenging problem. To circumvent this issue, researcher suggest to operate in lower dimensional latent space (latent diffusion models, LDMs) \cite{rombach2022high}, or generate low-resolution images and then upscale them with super-resolution diffusion models (cascaded diffusion model, CDM) \cite{ho2022cascaded} or mixtures-of-denoising-experts (eDiff-I) \cite{balaji2022ediffi}. Recently, end-to-end high resolution image generation have been proposed, either by improving the training loss \cite{hoogeboom2023simple} or by generating multiple resolutions simultaneously \cite{gu2023matryoshka}.

The aforementioned solutions involve require from-scratch training or fine-tuning, requiring additional computation cost. In this paper, we introduce a novel technique, \textit{upsample guidance}, which enables higher resolution sampling without any additional training or external models, but simply by adding a single term involving minimal computation. 

As shown in \cref{fig:showcase}, upsample guidance can be universally introduced to any types of diffusion model, including pixel-space, latent-space, or even video diffusion model. Moreover, it is fully compatible with any diffusion models or all previously proposed techniques that improve or control diffusion models such as SDEdit \cite{meng2021sdedit}, ControlNet \cite{zhang2023adding}, LoRA \cite{hu2021lora, simo2022lora}, and IP-Adapter \cite{ye2023ip}. 
Surprisingly, our method can even allows to generate higher resolution images that never shown in the training dataset, such as $64^2$ resolution images of CIFAR-10 dataset \cite{krizhevsky2009learning}, which has $32^2$ resolution images.

We demonstrate the results of applying upsample suidance across numerous pre-trained diffusion models and compare these with cases where our method was not applied. Additionally, we show the feasibility of spatial and temporal upsampling in video generation models. Finally, we conduct experiments on the selection of an appropriate guidance scale.

\section{Related Works}

Various ideas have been proposed for generating high-resolution samples using diffusion models. However, many of these require modifications to the architecture or traning from scratch. Here, we focus on a method that leverages pre-trained models to generate at resolutions higher than their trained resolution. 

\subsection{Super-Resolution}
\label{sec:superresolution}
An intuitive solution is to use pre-trained models to generate low-resolution samples and then upscale them to high resolution through a super-resolution model. Cascaded diffusion models (CDM) perform super-resolution using diffusion models that take low-resolution images as a condition \cite{ho2022cascaded} . This method has been applied to high-performance text-to-image models such as IMAGEN \cite{ho2022imagen} and DeepFloyd-IF \cite{sd2023IF}. However, this approach involves muptiple diffusion models and sampling process, requiering additional training and heavy computational cost.

A similar method in practice involves upscaling an image generated by a diffusion model using a relatively lightweight super-resolution model, followed by applying SDEdit \cite{meng2021sdedit} with the same diffusion model to enhance details in the high-resolution image. This technique is implemented under the name "HiRes.fix" in a well-known web-based diffusion model UI \cite{automatic2022webui}. However, a drawback is the additional encoding and decoding operations required when it is combined with LDMs.

\subsection{Fine-Tuning}

Even models trained at a fixed low resolution can generate higher resolution images better when fine-tuned on datasets with higher resolutions and various aspect ratios \cite{zheng2023any}. For instance, although the Stable Diffusion v1.5 model \cite{rombach2022high} was trained at a $512^2$ resolution, several models fine-tuned near a $768^2$ resolution are widely used. However, as the resolution increases, the computational cost required for fine-tuning also rises considerably, making it challenging to train on higher resolutions. 

\section{Background}
In this section, we introduce the fundamental concepts of diffusion models and guidance, which are essential for understanding our method.

\subsection{Diffusion Models} 

Diffusion models transform an original sample $x_0$ from the dataset into a noised sample $x_t$ through a forward diffusion process, eventually reaching pure noise $x_T$ that can be easily sampled. Many diffusion models follow the formalization of denoising diffusion probabilistic models (DDPMs) \citep{ho2020denoising} that use Gaussian noise. Specifically, the noised sample is given by:
\begin{equation}
\label{eq:forward_process}
    x_t=\sqrt{\alpha_t} x_0 + \sqrt{1-\alpha_t} \epsilon_t,
\end{equation}
which represents a linear combination of the signal $x_0$ and noise $\epsilon_t \sim \mathcal{N}(0, I)$. The term $\alpha_t$ is a noise schedule that affects the signal-to-noise ratio $\textrm{SNR} = \frac{\alpha_t}{1 - \alpha_t}$ and monotonically decreases with respect to time $t$.

The generation in DDPMs corresponds to a backward diffusion process, starting from $x_T$ and approximately sampling the distribution $p(x_{t-1}|x_t)$. The expected value of this conditional distribution is estimated by a noise predictor $\epsilon(x_t,t)$, which takes the noised sample $x_t$ and time $t$ as inputs to predict $\epsilon_t$.  
Note that the noise predictor has a U-Net architecture \cite{ronneberger2015u}, allowing it to accept inputs at resolutions other than the trained resolution. This flexibility underscores the adaptability of the model to handle a variety of image sizes.

\subsection{Guidances for Diffusion Models}

Techniques have been proposed for conditionally sampling images corresponding to specific classes or text prompts by adding a guidance term to the predicted noise. \citet{ho2022classifier} added the gradient of the log probability predicted by a classifier to $\epsilon(x_t, t)$, enabling an unconditional diffusion model to generate class-conditioned images. Subsequently, classifier-free guidance (CFG) was proposed. Instead of using a classifier, the noise predictor's architecture was modified to accept condition $c$ as an input. The following formula is then used as the predicted noise: 
\begin{equation}
\label{eq:cfg}
    \tilde{\epsilon}(x_t,t;c)=
    {\underbrace{\epsilon(x_t, t)}_{\text{denoise}}}+
    w{\underbrace{[\epsilon(x_t, t; c) - \epsilon(x_t, t)]}_{\text{guidance}}}.
\end{equation}
Here, $w$ represents the guidance scale. It has been commonly observed that proper adjustment of the scale improves the alignment with the condition and the fidelity of the generated images. 

\section{Method}
In this section, we first introduce signal-to-noise ratio (SNR) matching, a core concept crucial for understanding our method, and then derive upsample guidance based on it. We also explain the considerations when applying it to LDMs that include an encoder-decoder structure. 

\subsection{SNR Matching}

\begin{figure}
\vskip 0.2in
\begin{center}
\centerline{\includegraphics[width=\columnwidth]{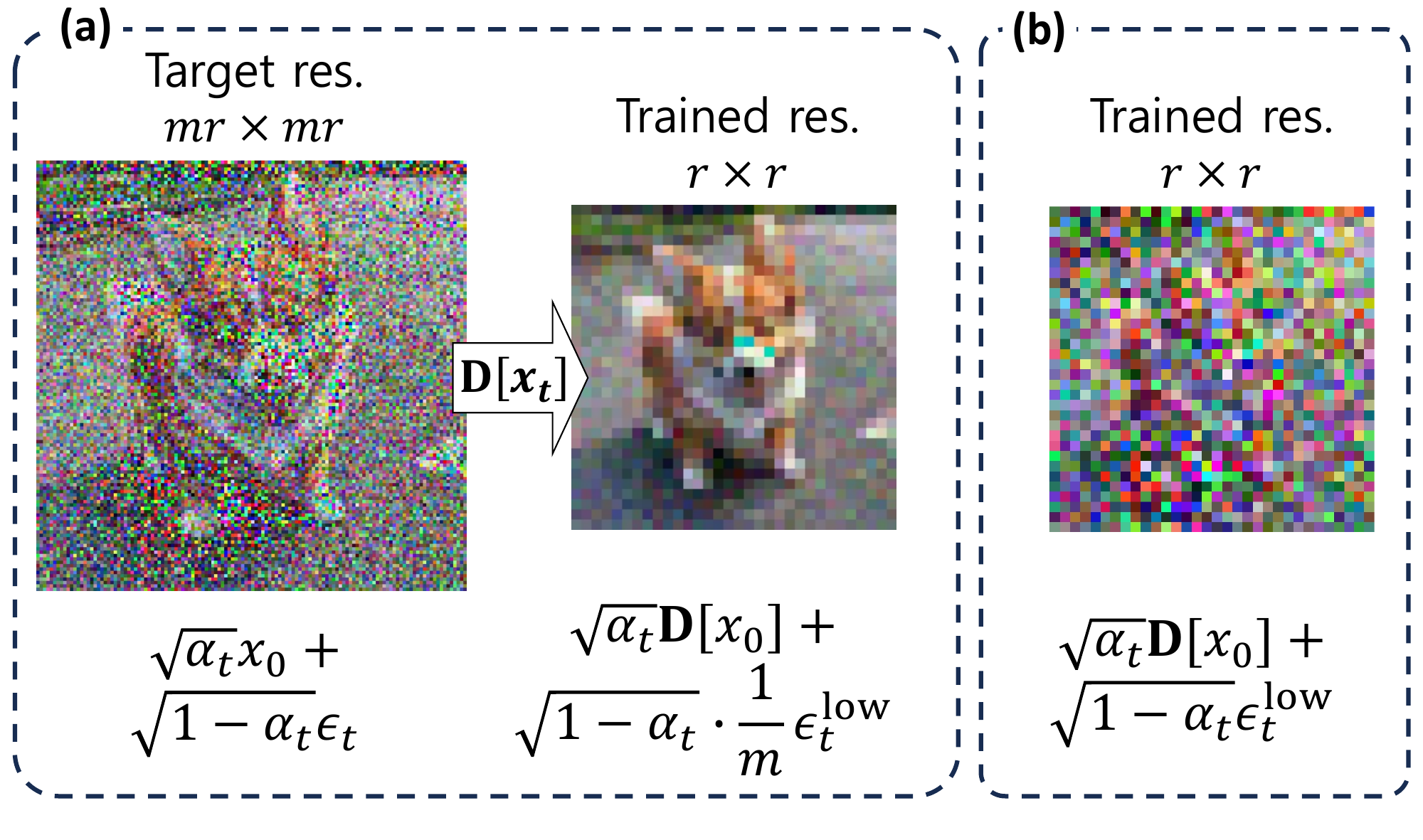}}
\caption{
Consistency between different resolutions.
(a) Downsampled image generated by the diffusion model at the target resolution. (b) Image generated at the trained resolution.
The noise reduction due to downsampling creates a significant difference in the recognizability between the central and right images at the trained resolution, indicating a change in their signal-to-noise ratio.
For this example, $\alpha_t=0.85$ is used.
}
\label{fig:xt}
\end{center}
\vskip -0.2in
\end{figure}

When a diffusion model is trained at a resolution of $r \times r$, consider the scenario of generating images at a target (high) resolution that is $m$ times higher, namely $mr \times mr$. If images can be ideally generated at all resolutions, it is reasonable to expect that the result of downsampling the target resolution by a scale of $1/m$ should resemble the outcome sampled at the trained (low) resolution. In this context, let's assume that the same model takes a downsampled image $\D[x_t]$ as input during the generation process. Here, $\D$ represents a downsampling operator that takes the average of $m\times m$ pixels.

As \cref{fig:xt} demonstrates, this downsampled image $\D[x_t]$ (the second image) significantly differs from $x^{\textrm{low}}_t$ (the third image) that was generated from the diffusion model trained at the low resolution.
More specifically, the trained image follows
\begin{equation}
    x^{\textrm{low}}_t = \sqrt{\alpha_t} \D [x_0] + \sqrt{1-\alpha_t} \epsilon^{\textrm{low}}_t,
\end{equation}
where $\epsilon^{\textrm{low}}_t$ represents the standard Gaussian noise with a size of trained resolution.
However, the downsampled image can be obtained from Equation \eqref{eq:forward_process} by applying the linear downsampling operator $\D$,
\begin{equation}
    \D[x_t] = \sqrt{\alpha_t} \D [x_0] + \frac{1}{m} \sqrt{1-\alpha_t} \epsilon^{\textrm{low}}_t.
\end{equation}
Here, the standard deviation of noise is reduced to $1/m$, because $\D$ averages every $m\times m$ pixels. This directly results form the central limit theorem.
Therefore, some adjustments are necessary to make $x^{\textrm{low}}_t$ and $\D[x_t]$ equivalent \cite{hwang2023resolution}.
This requires matching both SNR and overall power:
\begin{align}
    &\mathrm{SNR}^{\mathrm{low}} = \frac{m\alpha_t}{1-\alpha_t} = m \cdot \mathrm{SNR}, \\
    &P = \alpha_t + \frac{1}{m^2} ( 1- \alpha_t).
\end{align}

Since the SNR is a function of time determined by the noise schedule $\alpha_t$, we can find the adjusted time $\tau$ such that $\mathrm{SNR}^{\mathrm{low}}(\tau)=m\cdot \mathrm{SNR}(t)$. Furthermore, by multiplying by $1/\sqrt{P}$, we can make the overall power equivalent to that of the target resolution. 
Therefore, the proper noise predictors of high and low resolutions are associated with time and power adjustments as follows
\begin{equation}
\label{eq:eps_low}
    \D [\epsilon^\mathrm{adj}(x_t,t)] = \frac{1}{m} \epsilon\left( \frac{1}{\sqrt P} \D[x_t], \tau \right),
\end{equation}
where the factor of $1/m$ is multiplied to adjust the variance of trained resolution to 1.

\subsection{Upsample Guidance}
\begin{figure}
\vskip 0.2in
\begin{center}
\centerline{\includegraphics[width=\columnwidth]{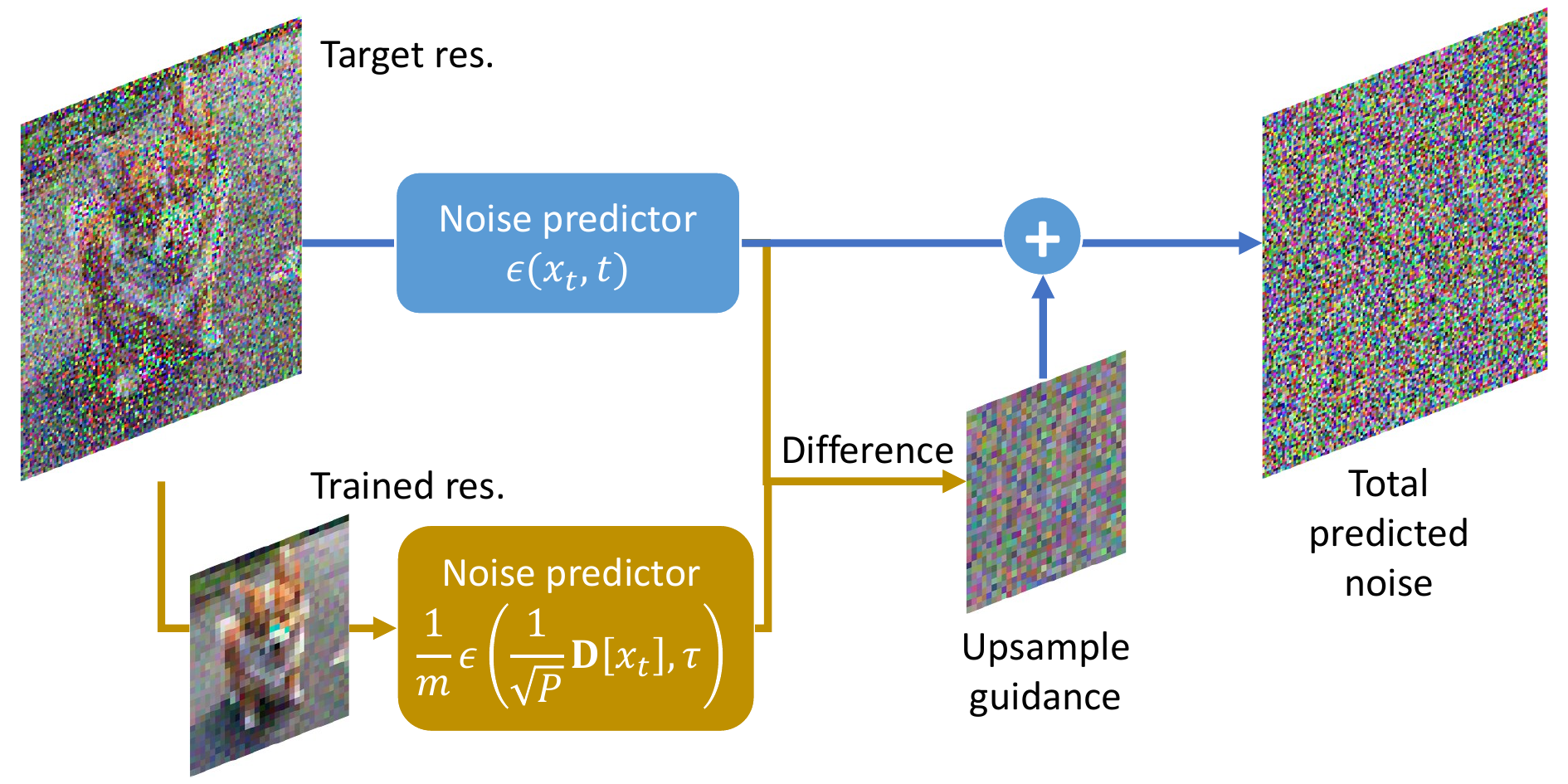}}
\caption{
Conceptual illustration of upsample guidance. The model receives the same noised images at two different resolutions in parallel, but time and power are adjusted at the trained resolution. The difference between the two predicted noises then acts as guidance, which is added to the total noise.}
\label{fig:method}
\end{center}
\vskip -0.2in
\end{figure}

Considering the consistency across diverse resolutions, we explore the decomposition of predicted noise at the target (high) resolution. Suppose that the target noise comprises both a component from the trained (low) resolution and its corresponding residual part:
\begin{equation}
    \epsilon(x_t,t) = \U\underbrace{\D[\epsilon(x_t,t)]}_{\textrm{trained  resolution}} + \underbrace{\{\epsilon(x_t,t) - \U\D[\epsilon(x_t,t)]\}}_\textrm{target resolution}.
\end{equation}
In this context, $\U$ represents the nearest upsampling operator with a scale factor of $m$, utilized to align dimensions between target and trained noise predictors.
The residual noise, $\epsilon - \U\D[\epsilon]$,  corresponds to the part that remains after removing the contribution of the low resolution.  

Now, recognizing the need for adjustments to ensure consistency among noise predictors at various resolutions, we substitute the term about trained resolution with the adjusted noise predictor in Equation \eqref{eq:eps_low}, as follows:
\begin{align}
    \epsilon^\mathrm{adj}(x_t,t) =& \U\underbrace{ \left[ \frac 1 m \epsilon\left( \frac{1}{\sqrt P} \D[x_t], \tau \right) \right]}_{\textrm{trained  resolution}} \nonumber +\\& \underbrace{\{\epsilon(x_t,t) - \U\D[\epsilon(x_t,t)]\}}_\textrm{target resolution}.
\end{align}
This model parallelly sees and predicts noises at both resolutions.

Finally, we consider interpolation between the naive sampling at the targe resolutin with the parallel sampling at the trained resolution,
\begin{align}
\label{eq:upsample_guidance}
    &\tilde\epsilon(x_t,t) =(1-w_t)\epsilon(x_t,t) + w_t\epsilon^\mathrm{adj}(x_t,t) =\nonumber \\
    & \epsilon(x_t,t) + w_t \underbrace{\U\left[ \frac{1}{m} \epsilon\left( \frac{1}{\sqrt P} \D[x_t], \tau \right) - \D[\epsilon(x_t,t)] \right]}_\textrm {upsample guidance}.
\end{align}
This structure resembles Equation \eqref{eq:cfg} and can be interpreted similarly as guidance. Consequently, we named it ``upsample guidance (UG),'' where $w_t$ functions as the guiding scale, which may generally depend on time.
Similar to how CFG incorporates the shift from unconditional to conditional noise, UG represents the influence pushing the model towards consistency with the trained low-resolution component.

\subsection{Adaptation on LDMs}

\begin{figure}
\vskip 0.2in
\begin{center}
\centerline{\includegraphics[width=\columnwidth]{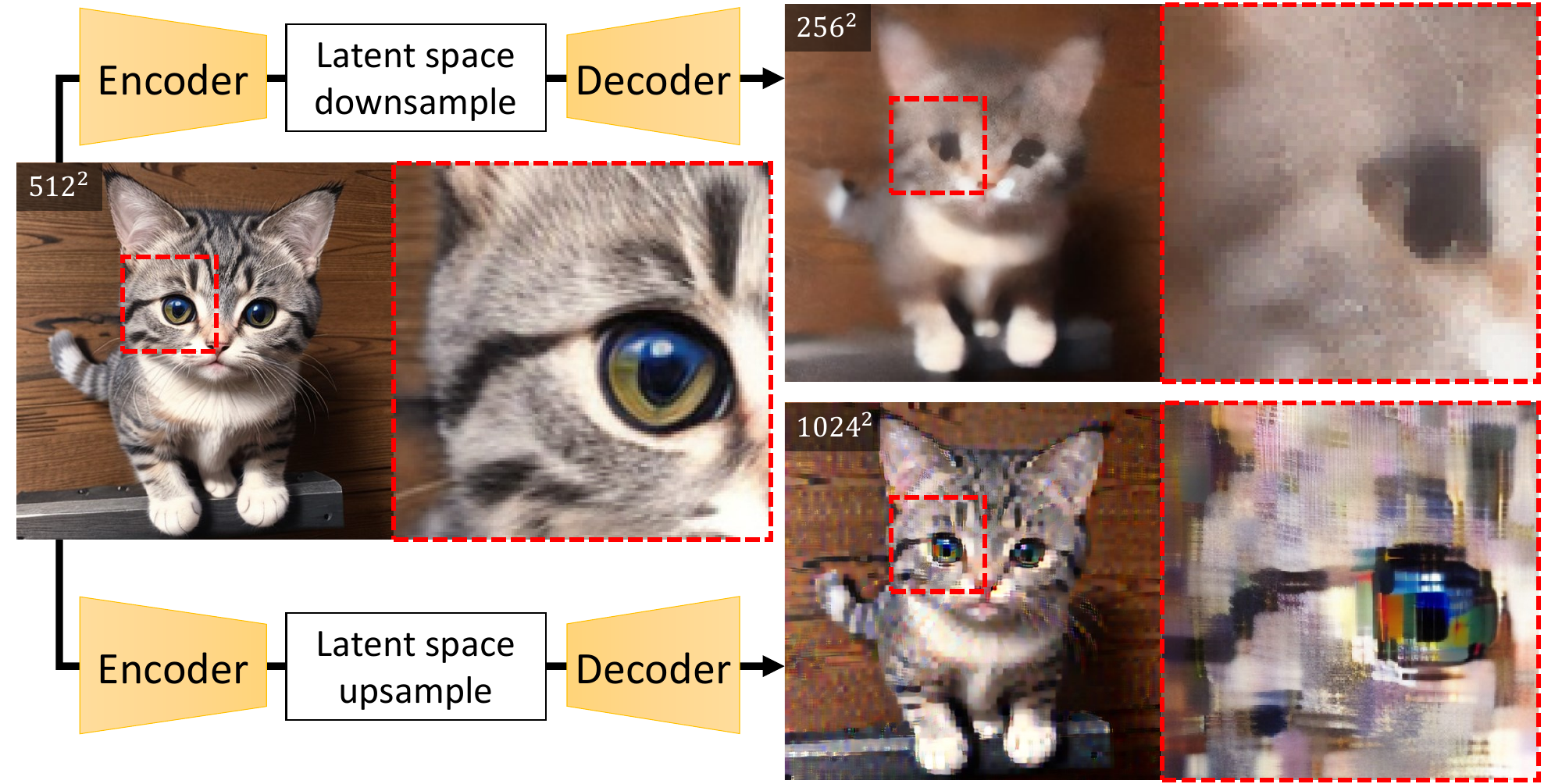}}
\caption{
Artifacts of encoder-decoder in a LDM. When an image is upsampled or downsampled in the latent space of an LDM and then decoded back into pixel space, artifacts are introduced. The variational autoencoder introduces nonlinearity in the implementation of upsample guidance, and significant degradation can be observed in both cases.}
\label{fig:artifact}
\end{center}
\vskip -0.2in
\end{figure}

\begin{figure*}[!ht]
\begin{center}
    \centerline{\includegraphics[width=0.9\textwidth ]{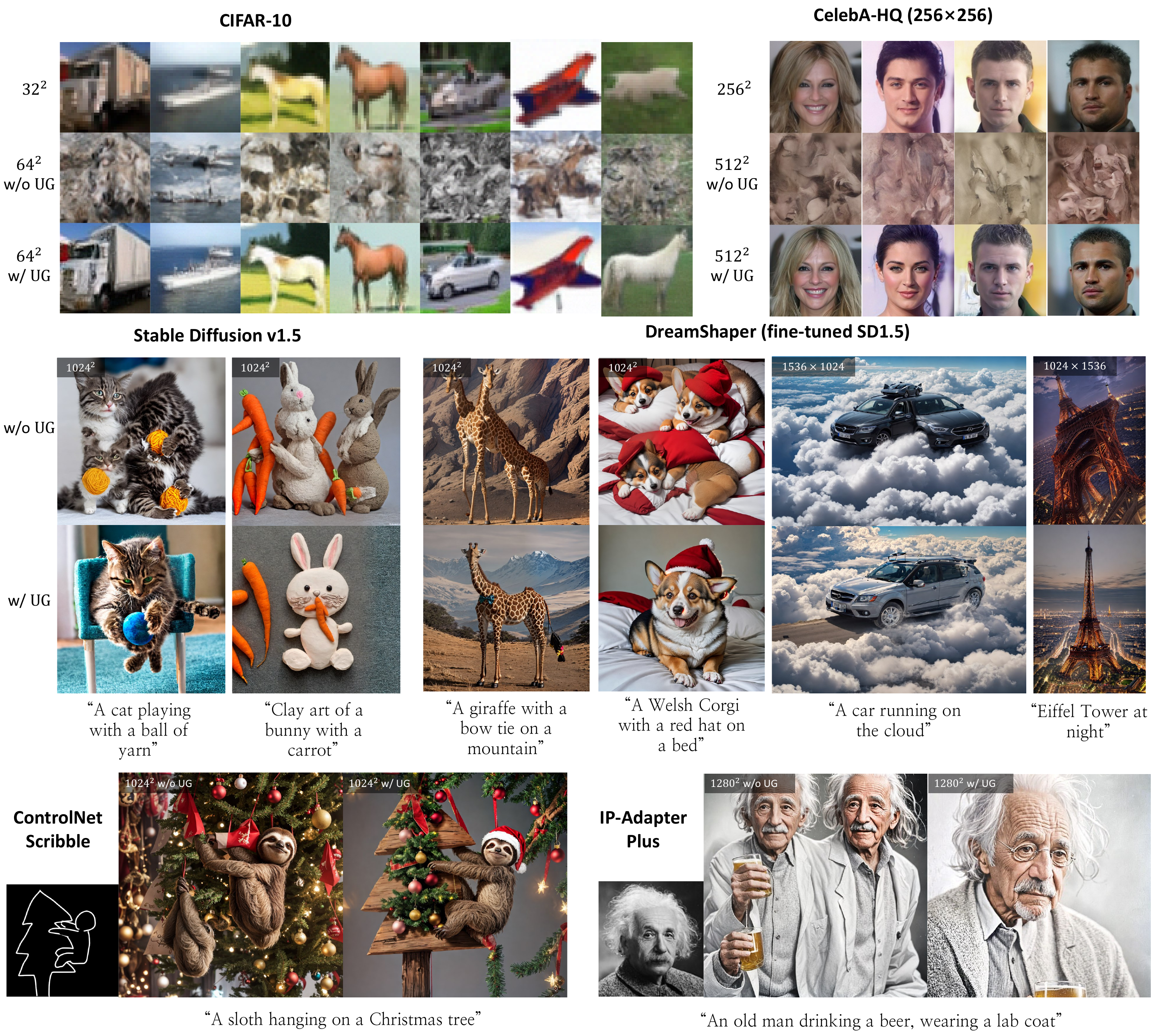}}
    \caption{Upsampling across various image generation models, resolutions, and conditional generation methods. Unconditional image generation, such as CIFAR-10 and CelebA-HQ, was sampled in the pixel space. For the text-to-image models, the left side of the images represents results without UG, while the right side shows results with UG. We used DreamShaper \cite{dreamshaper2023} as an example of fine-tuned LDM. The paired images are all generated from the same initial noise. Across different models, resolutions, prompts, and conditioning, consistently better images were obtained with UG. Notably, our method effectively resolved artifacts where multiple subjects were generated or bad anatomy was present.}
    \label{fig:images}
\end{center}
\vskip -0.2in    
\end{figure*}

The aforementioned derivation heavily relies on the linearity of operators $\U$ and $\D$. Nonetheless, in the context of LDMs, the pixel space undergoes a transformation into the latent space using a nonlinear variational autoencoder (VAE) \cite{kingma2013auto}. Consequently, it is crucial to proceed with caution, as the latent space of a downsampled image at the target resolution may not align with the latent space of the resolution it was originally trained on. The outcomes of downsampling in latent space and subsequently decoding back into pixel space are shown in \cref{fig:artifact}.

However, we discovered a viable solution to this challenge by heuristically tailoring $w_t$ to be time-dependent. Specifically, we designed $w_t$ to decrease or be set to zero when $t$ is close to zero, preventing the upsample guidance from introducing artifacts. While various designs for $w_t$ are conceivable, in Section \ref{sec:kt_analysis}, we introduce the most straightforward parameterized design using the Heaviside step function $H$ to investigate the influence of scale magnitude $\theta$ and time threshold $\eta$. The formulation is expressed as follows:

\begin{equation}
\label{eq:kt}
    w_t = \theta \cdot H(t-(1-\eta) T).
\end{equation}

\section{Experiments}

The core concept behind upsample guidance lies in the SNR matching during the downsampling process. As a result, it can be extended to diverse data generation tasks, not confined to images alone. Moreover, its compatibility extends to any pre-trained model, conditional generation, and application techniques. In this section, we showcase the outcomes when applied to various image generation models and applications. Specifically, we explore spatial and temporal upsampling in video generation. Subsequently, we conduct an ablation study to evaluate scenarios where the adjustments proposed in \cref{eq:upsample_guidance} are not implemented. Lastly, a quantitative analysis on the guidance scale is performed to help the design of $w_t$.

\subsection{Image Upsampling}

As upsample guidance requires only a straightforward linear operation on the predicted noise, it exhibits compatibility with a wide array of models and applications.
In our study, we used a pre-trained unconditional model trained on CIFAR-10 and CelebA-HQ $256^2$ \cite{karras2018progressive} datasets to generate images at twice the resolution using a constant $w_t$. We also sampled using UG with $m=2$ for text-to-image models based on stable diffusion v1-5, and checked its capability on a fine-tuned model, different aspect ratios and image conditioning techniques.

\cref{fig:images} presents images that are slightly cherry-picked to aid in understanding the impact of upsample guidance. Each image pair contains images generated from the same initial noise. For images with different resolutions, initial noise was resized and its variance was adjusted accordingly. Upon carefully examining some samples from CIFAR-10, UG sometimes alters coarse contents (overall colors and shapes) between $32^2$ and $64^2$ resolutions, with details emerging at higher resolutions that were not present at lower ones. This suggests that UG does more than just interpolation or sharpening; it actually generates new meaningful features. 
For the comparison between low and high resolution in LDMs and more extensive non-cherry-picked samples, please refer to \cref{apdx:more_image_generation}.

\begin{figure}
\begin{center}
    \centerline{\includegraphics[width=0.8\columnwidth ]{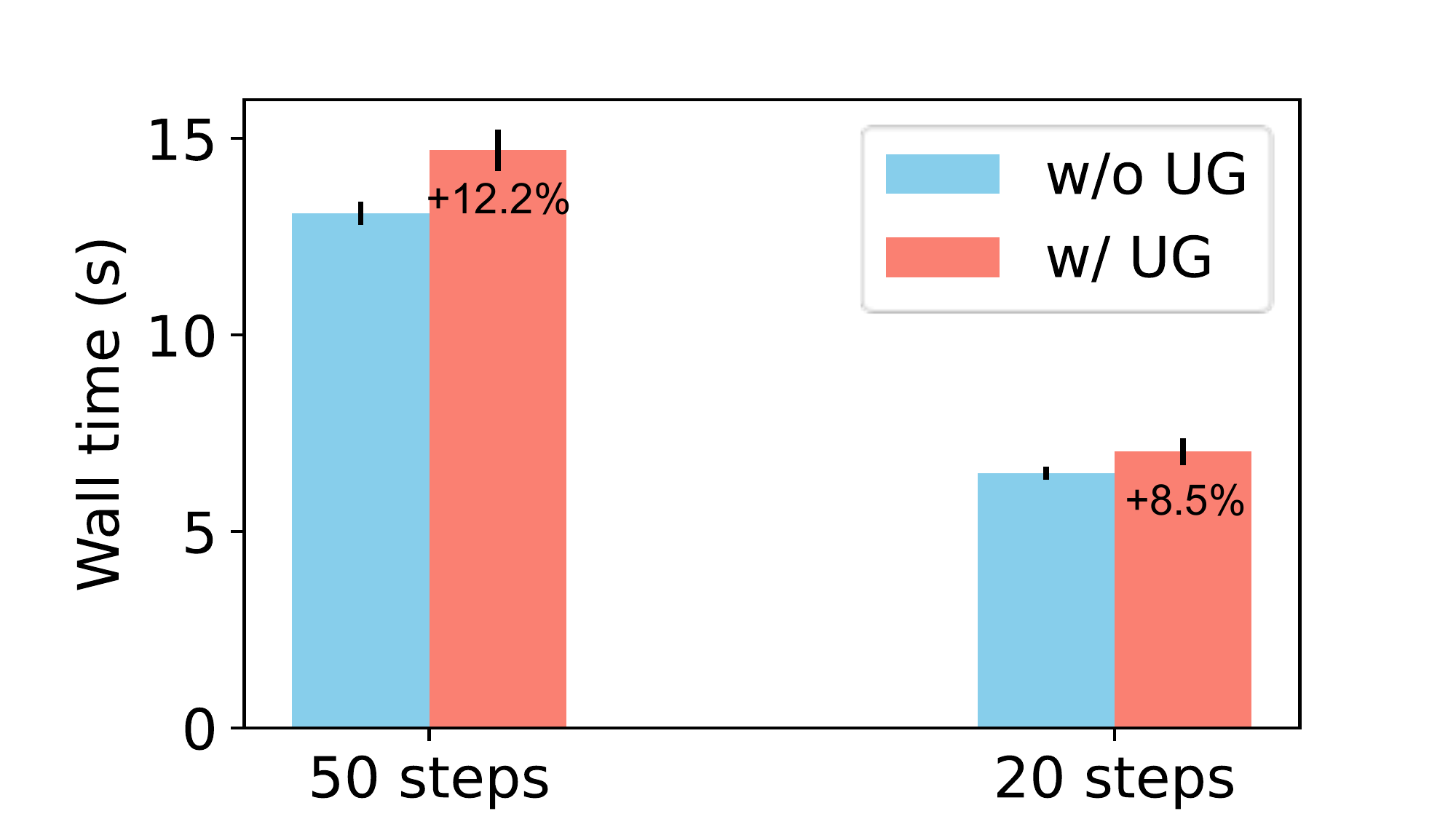}}
    \caption{
    Computational cost comparison for upsample guidance (UG). Wall time for computation is compared with and without the use of UG. The percentages on the bars indicate the proportion of additional time attributed to UG.}
    \label{fig:walltime}
\end{center}
\vskip -0.2in    
\end{figure}

We roughly measured the additional computational time introduced by UG. In Stable Diffusion v1-5, using an RTX3090 GPU at $1024^2$ resolution with a scale factor of $m=2$ and $\eta=0.5$, we measured the wall time from sampling in the latent space to converting into an RGB image, as shown in the \cref{fig:walltime}. 
The extra computation for UG is minimal, given that the dimension of the noise prediction $\epsilon(\frac{1}{\sqrt P}\D[x_t], \tau)$ at trained resolution is $1/m^2$ times the dimension of $\epsilon(x_t,t)$. Additionally, this supplementary computation is only applied when $t\geq(1-\eta)T$.
Therefore, the cost is less than $\eta / m^2 = 1/8$ of the naive sampling time. However, in LDMs, decoding also consumes the time, so the portion of cost due to UG decreases as the sampling step becomes shorter. With the recent advancements in sampling methods \cite{song2020denoising,  liu2022pseudo, luo2023latent} leading to a reduction in the number of inference steps, our method becomes more competitive, requiring only $\leq10\%$ additional computation cost within 20 inference steps.

\subsection{Video Upsampling}
\begin{figure*}[!ht]
\begin{center}
    \centerline{\includegraphics[trim=0cm 1.5cm 0cm 2cm, clip, width=\textwidth ]{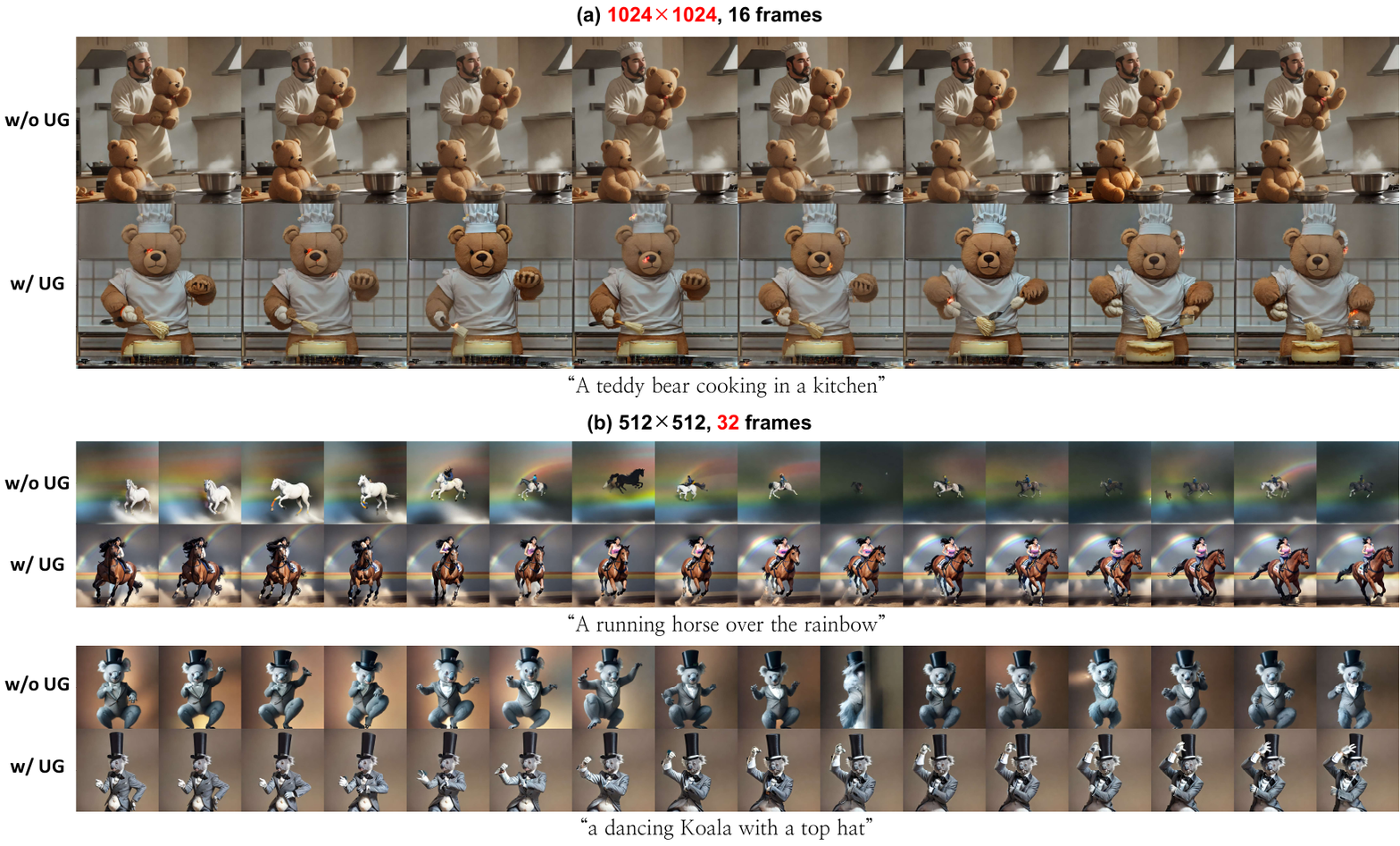}}
    \caption{
    Spatial and temporal video upsampling. Frames of videos are generated using AnimateDiff with UG applied. (a) Spatial upsampling by a factor of 2, similar to images. (b) Temporal upsampling with the number of frames upsampled by a factor of 2.
    Note that for visibility, only odd-numbered frames from each sequence are displayed.
    }
    \label{fig:animatediff}
\end{center}
\vskip -0.2in    
\end{figure*}

Upsample guidance can also enhance video upsampling by addressing both spatial and temporal resolution. To illustrate this, we employ AnimateDiff \cite{guo2023animatediff}, a video generation model that integrates a motion module into a text-to-image model.
In AnimateDiff, a video is represented as a sequence of color, time, width, and height in latent space, basically a tensor with the shape [C, T, W, H]. While we can upsample in the spatial dimensions [W, H] as above, it's also possible to upsample in the temporal dimension T, increasing the number of frames by a factor of $m$. Assuming UG gives robustness for temporal resolution, we expect an increase in frames per second rather than an extension of time length, similar to the case with images. 

\cref{fig:animatediff} shows the results of applying UG across these two dimensions. For spatial upsampling, issues like multiple subjects appearing and misalignment with text prompts were resolved thanks to UG, indicating that spatial UG works similarly in video generation as it does for images.

For temporal upsampling, we kept the spatial size constant and generated 32 frames, double the 16 frames AnimateDiff was trained on. Without UG, there was a complete failure in maintaining temporal consistency, and sometimes even adjacent frames lost continuity. However, with UG, the videos were overall consistent at a level similar to the trained temporal resolution, and greater continuity was also appeared in the subject's movements. This difference is more pronounced when viewing the videos in playback rather than as listed frames. 

\subsection{Ablation Study on Time and Power Adjustments}
\begin{figure}
\begin{center}
    \centerline{\includegraphics[width=\columnwidth ]{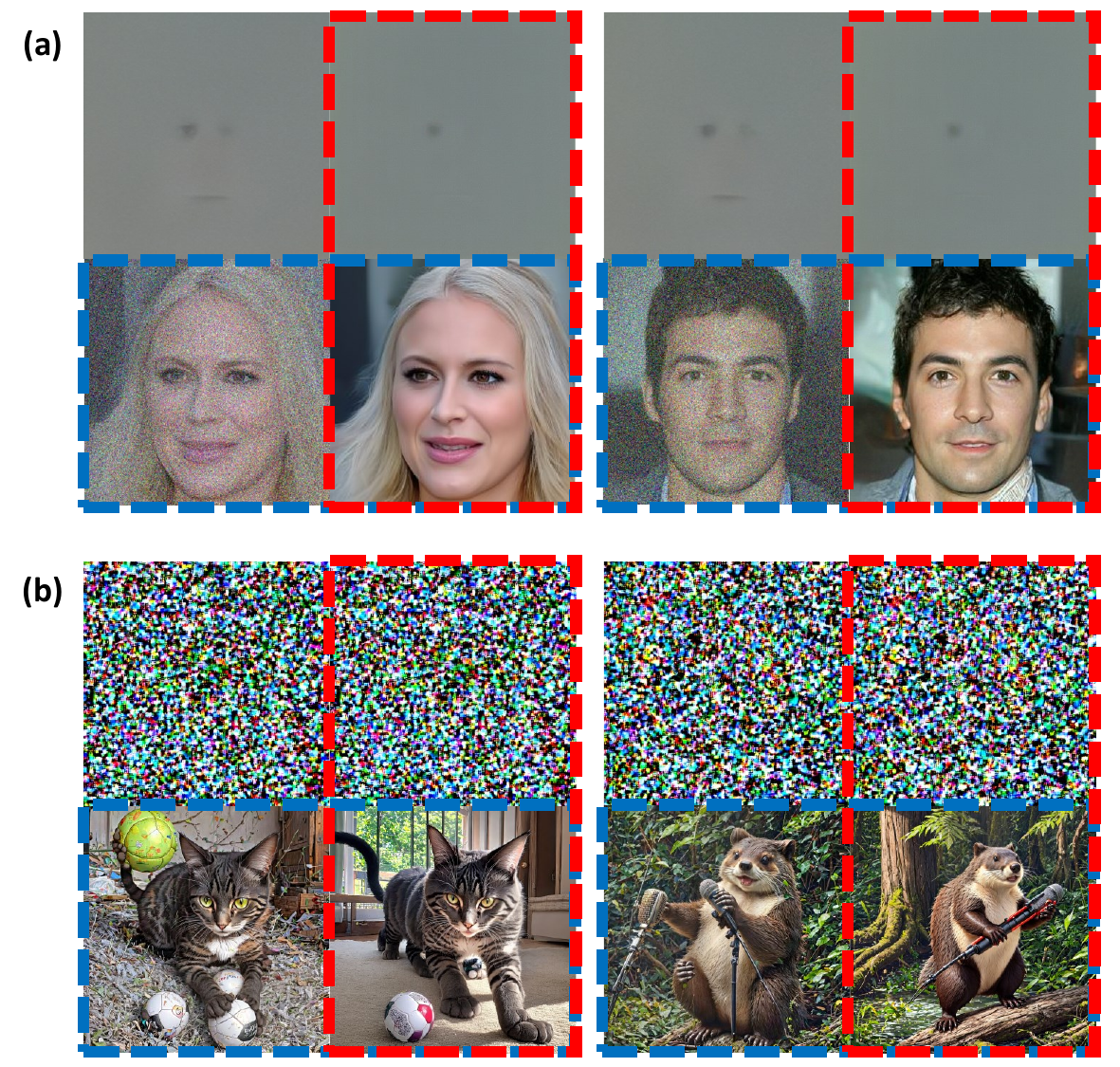}}
    \caption{ 
Effects of time and power adjustments in UG is indicated by red and blue dashed boxes, respectively. Two images are generated from (a) the CelebA-HQ model and (b) the text-to-image model, respectively, with and without either time adjustment, power adjustment, or both.}
    \label{fig:ablation}
\end{center}
\vskip -0.2in    
\end{figure}

So far, we have seen that our method effectively suppresses artifacts that could occur at higher resolutions. However, some might question the necessity of the time and power adjustment presented in \cref{eq:eps_low}. Therefore, we illustrate here that each adjustment is indeed essential, and how the images ruined when either one or both adjustments are not made.

As illustrated in \cref{fig:ablation}, both adjustments are essential, and the absence of each leads to image degradation in different ways. Without $\tau$ in \cref{eq:eps_low} (outside the red dashed boxes), the model fails in denoising due to the mismatch between the learned SNR at the time and the SNR of the input noised sample, resulting in residual noise. Without $1/\sqrt P$ (outside the blue dashed boxes), the model confronts samples with variances it has never learned, resulting in complete failure. This suggests that the noise predictor $\epsilon(x_t,t)$ is highly sensitive to time, variance, and SNR, making our method crucial. 

\subsection{Analysis on Guidance Scale}
\label{sec:kt_analysis}

We observed that for diffusion models in pixel space, it is acceptable to keep the guidance scale constant, but for LDMs, the guidance scale needs to be reduced near $t=0$ to eliminate artifacts as shown in \cref{fig:artifact}. To quantitatively analyze the impact of the guidance scale, we measured changes in LDMs using a time-independent $w_t$ and by parameterizing it as in \cref{eq:kt}.  

\subsubsection{Sampling on Pixel Space}
\label{sec:sampling_on_pixel_space}
\begin{figure}
\begin{center}
    \centerline{\includegraphics[width=0.8\columnwidth ]{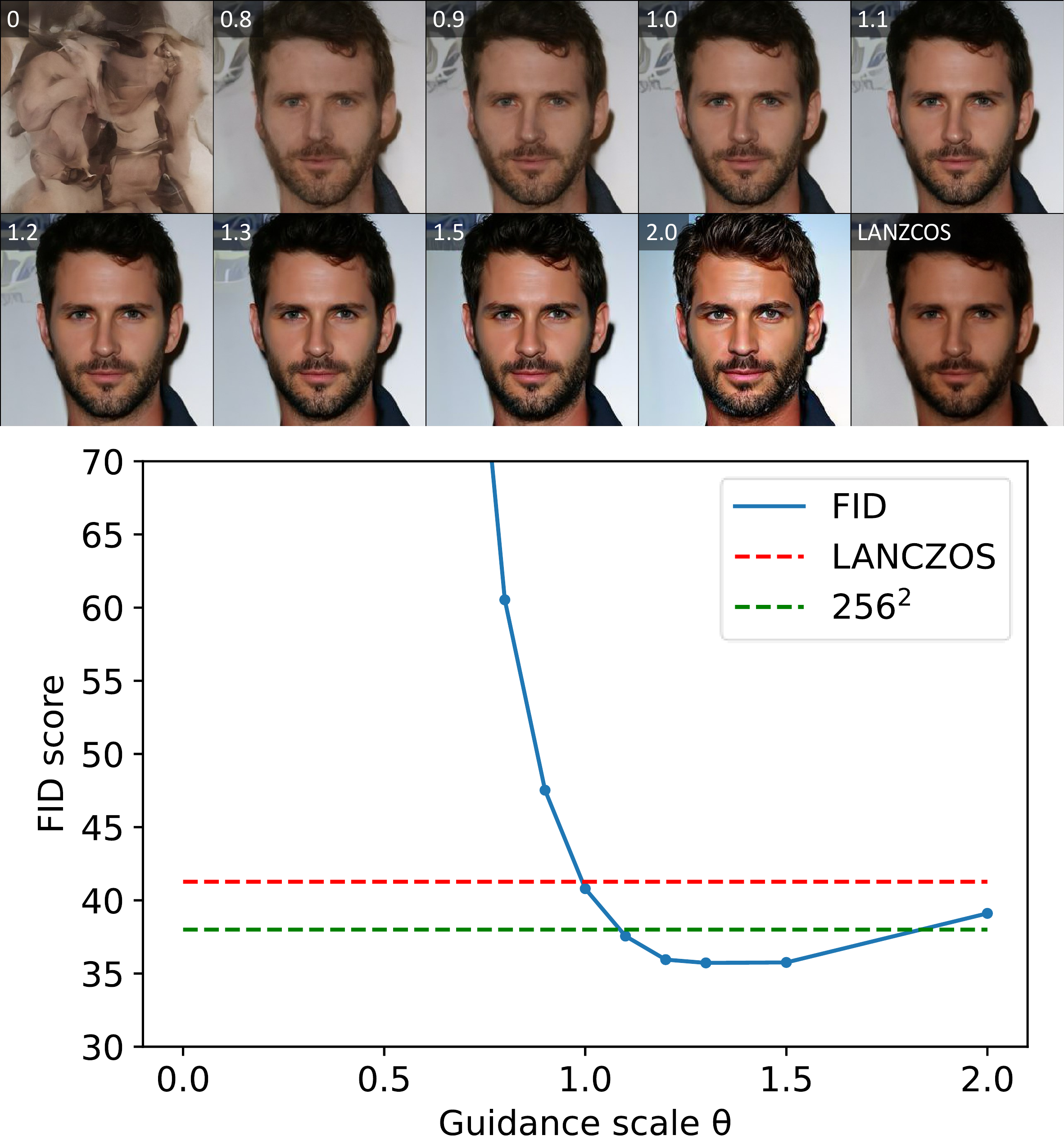}}
    \caption{Fidelity of generated images across different guidance scale $w_t$ (numbers in upper left corners) measured by the FID score (lower is better).  The label ``LANCZOS'' refers to the images originally generated at a size of $256^2$ by the model and then upsampled to $512^2$ using Lanczos resampling. The green dashed line represents the FID between images generated by the model at $256^2$ trained resolution and the CelebA-HQ $256^2$ dataset. }
    \label{fig:celeba}
\end{center}
\vskip -0.2in    
\end{figure}

We empirically found that for pixel space diffusion models, keeping the guidance scale constant is effective. Thus, we recommend using a constant $w_t$, with $\eta=1$ and varying $\theta$ only in Equation \eqref{eq:kt}. After generating $512^2$ resolution images from a model trained on the CelebA-HQ $256^2$ using UG, we measured the fidelity via Fréchet inception distance (FID) \cite{heusel2017gans} to CelebA-HQ $512^2$. Results showed that as the guidance scale increased, the features and contrast became clearer. Astonishingly, at the optimal point ($\theta \approx 1.3$), the model outperformed not only resized images from the trained resolution but also achieved better fidelity compared to the dataset of the originally trained size of $256^2$, demonstrating that UG serves a role beyond simple interpolation or sharpening. 

\subsubsection{LDMs and Text-to-Image}
\label{sec:ldm}
\begin{figure}
\begin{center}
    \centerline{\includegraphics[width=\columnwidth ]{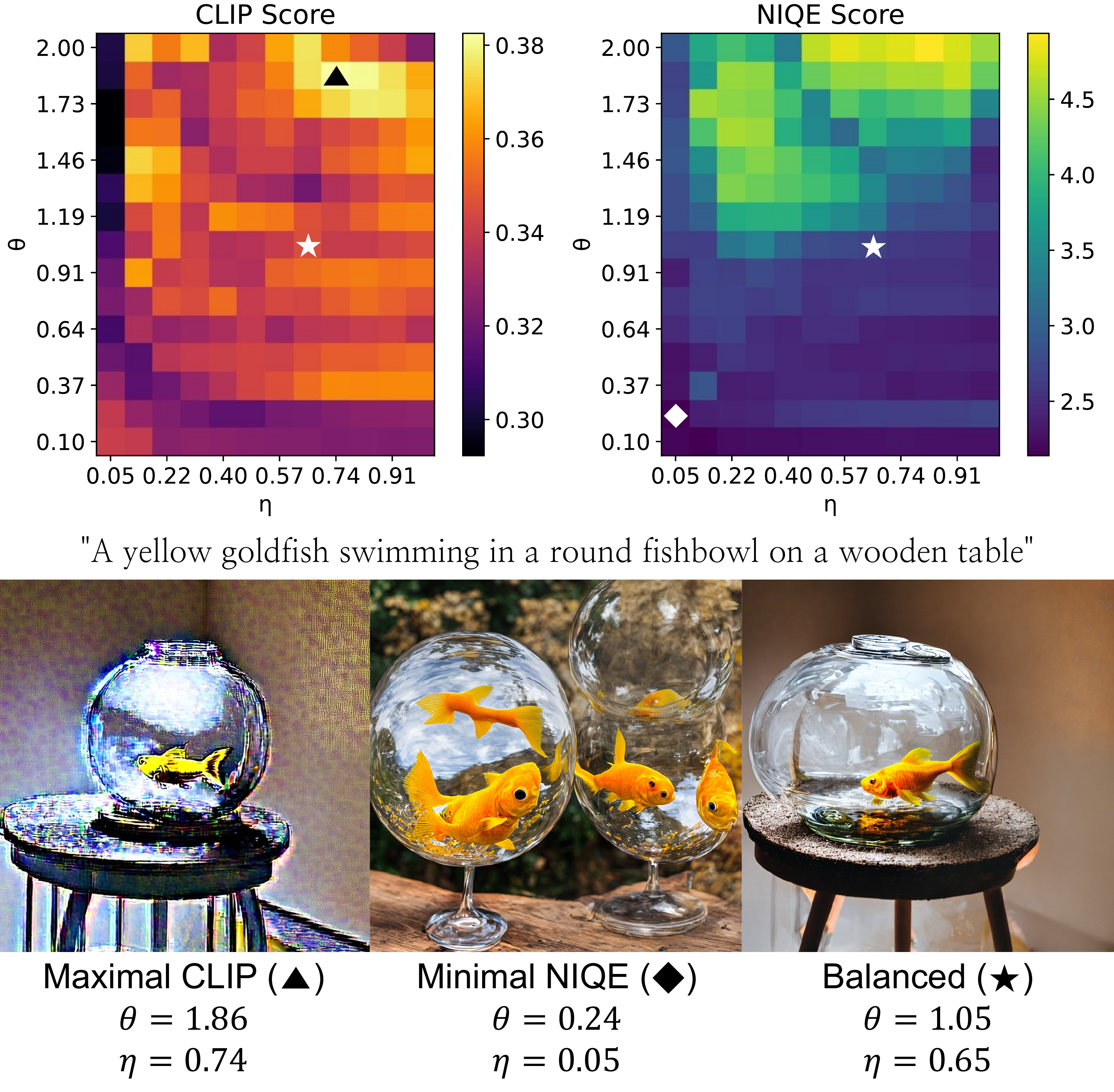}}
    \caption{Impact of guidance scale with CLIP score and NIQE metric. Specific image examples at three parameter points are selected as Maximal CLIP (triangle), Minimal NIQE (diamond), and Balanced (star). More Images are presented in \cref{apdx:grid}.}
    \label{fig:clip_niqe}
\end{center}
\vskip -0.2in    
\end{figure}

For LDMs, it's crucial to reduce the guidance scale to zero during the mid-stages of sampling to prevent artifacts as shown in \cref{fig:artifact}. However, if we tolerate the artifacts, the coarse structure of images can be better aligned with the text prompt. Therefore, choosing the guidance scale involves a trade-off between prompt alignment and image quality. 
To evaluate alignment and quality, we used CLIP score \cite{radford2021learning} and naturalness image quality evaluator (NIQE) \cite{mittal2012making} respectively.

As shown in \cref{fig:clip_niqe}, the CLIP score tends to increase with stronger guidance, indicating better alignment with the prompt. Conversely, NIQE scores worsen with strong guidance. At the optimal point for CLIP, image's coarse features align well with the prompt but lose photorealism. At NIQE's optimal point, the image appears locally natural and realistic but deviate significantly from the prompt. This trend was consistent across samples and prompts. We heuristically recommend $(\theta, \eta)\approx (1, 0.6)$ as a balanced setting.

\section{Conclusion}
In conclusion, we introduced upsample guidance, a training-free technique enabling the generation of high-fidelity images at high resolutions not originally trained on, demonstrating its applicability across various models and applications. Our method, derived from the diffusion process and not dependent on architecture, holds synergistic potential with any other techniques for high-resolution image generation. Moreover, UG uniquely enables the creation of images for datasets like CIFAR-10 $64^2$, where high-resolution data may not originally exist. 

In our experiments, we used a simple design of guidance scale for clarity, but there's room for enhancement through replacing it with more elaborated functions. While focusing on spatial upsampling, further exploration into the best practice for temporal upsampling in video and audio models is needed. Especially, for audio, careful implementation is necessary as temporal downsampling may shift pitch. 

The computational cost of UG is marginal, and ongoing research aimed at reducing inference steps further minimize the portion of time consumption due to UG in LDMs. We consider our method a universally beneficial add-on for generating high-resolution samples due to its ease of implementation and cost-effectiveness. 

\section{Impact Statements}

This paper presents work whose goal is to advance the field of Machine Learning. There are many potential societal consequences of our work, none which we feel must be specifically highlighted here.

\nocite{langley00}

\bibliography{upsample_guidance}
\bibliographystyle{icml2024}

\newpage
\appendix
\onecolumn
\section{Calculation of Adjusted Time $\tau$}

\begin{figure}
\begin{center}
    \centerline{\includegraphics[width=0.5\columnwidth ]{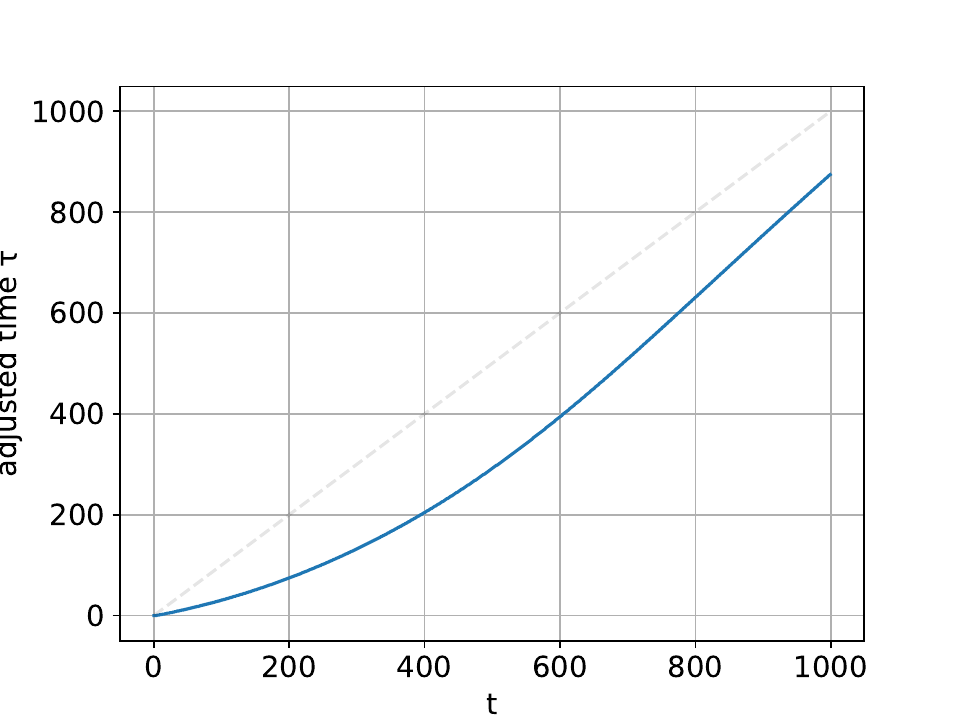}}
    \caption{Time adjstment with scale factor $m=2$ for the noise schedule of Stable Diffusion v1.5.}
    \label{fig:tau}
\end{center}
\vskip -0.2in    
\end{figure}

Time adjustment is for matching SNR, so analytically obtaining tau is possible by finding the inverse function of SNR over time. However, as most implementations encode time as an integer, it's sufficient to numerically approximate values rather than compute exact ones. Below is a Python code for numerically calculating time adjustment for integer times, and \cref{fig:tau} visualizes results calculated from a real model's $\alpha_t$.  

\begin{python}
def getTau(m, alphas):
    # m : scale factor
    # alphas : list of alpha_t
    snr = alphas / (1 - alphas)
    snr_low = alphas / (1 - alphas) * m**2
    log_snr, log_snr_low = np.log(snr), np.log(snr_low)
    def getSingleMatch(t):
        differences = np.abs(log_snr_low[t] - log_snr)
        tau = np.argmin(differences)
        return tau
    return [getSingleMatch(t) for t in range(len(alphas))]
\end{python}

 \newpage
\section{More Upsampling Examples}
\label{apdx:more_image_generation}
\begin{figure}[!hb]
\begin{center}
    \centerline{\includegraphics[width=0.9\columnwidth ]{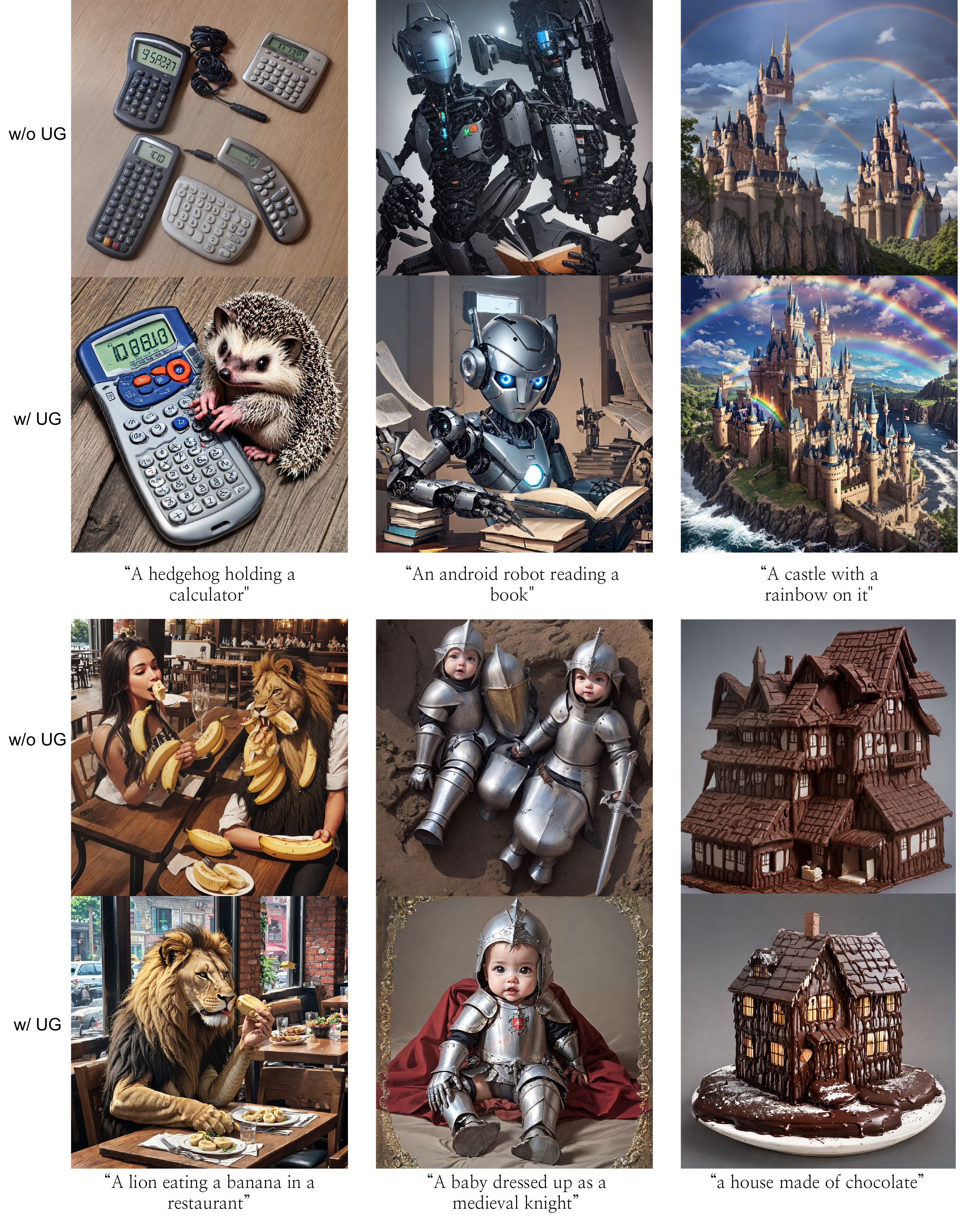}}
    \caption{Images generated with and without UG, sampled on $1280^2$ resolutions with $m=2$.}
\end{center}
\vskip -0.2in    
\end{figure}

\newpage
\begin{figure}[!hb]
\begin{center}
    \centerline{\includegraphics[width=0.68\columnwidth ]{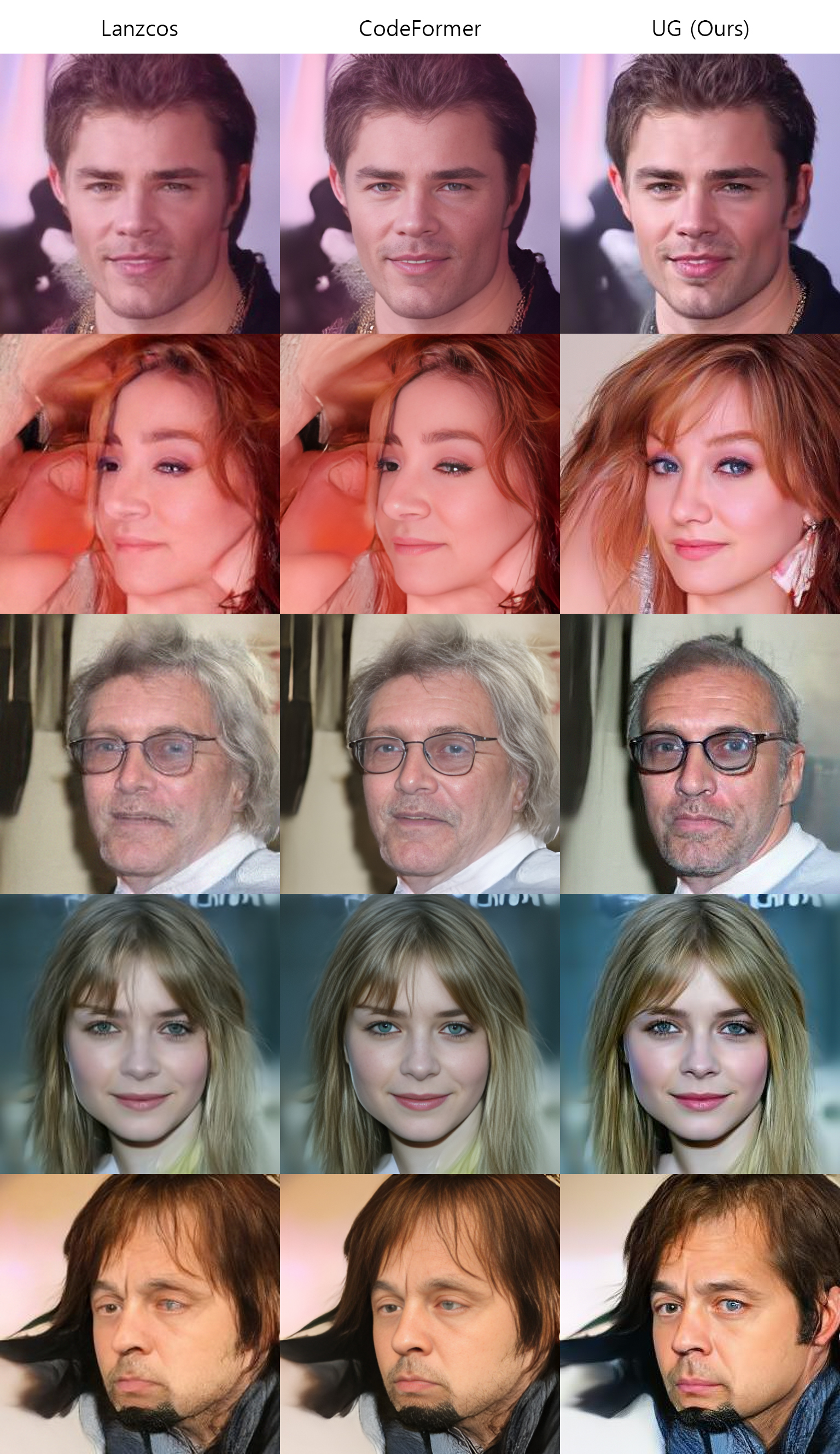}}
    \caption{Comparison of samples generated at a resolution of $512^2$ using different upscaling techniques, based on a model trained on the CelebA dataset at a resolution of $256^2$. `Lanczos' and `CodeFormer \cite{zhou2022codeformer}' refer to images upscaled from $256^2$ using their respective methods, while `UG' denotes samples generated with $w_t=1.35$, optimally chosen based on the experiments described in \cref{sec:sampling_on_pixel_space}}
\end{center}
\vskip -0.2in    
\end{figure}

\newpage
\begin{figure}[!hb]
\begin{center}
    \centerline{\includegraphics[width=0.75\columnwidth ]{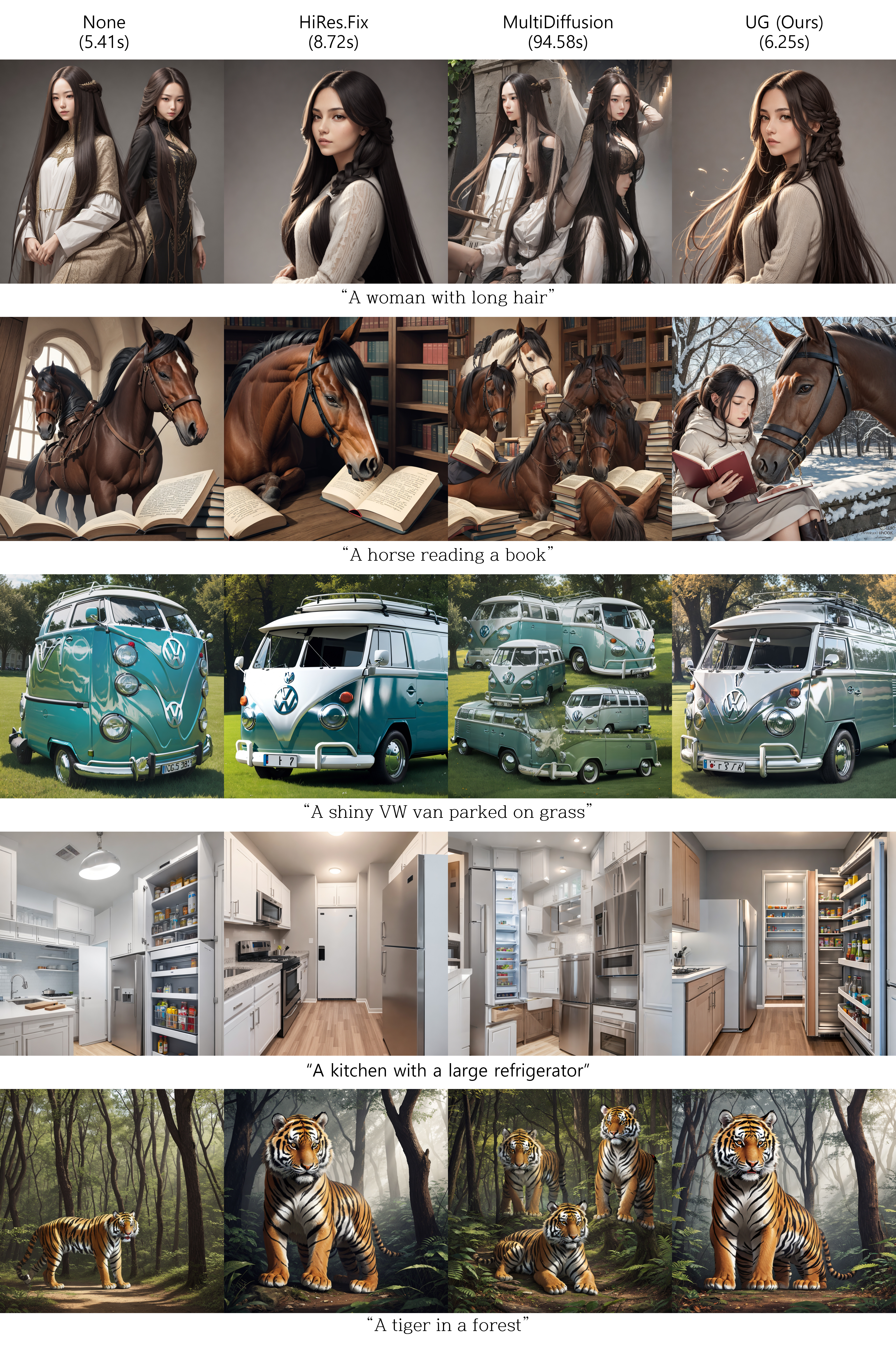}}
    \caption{Samples generated at a resolution of $1024^2$ using different upscaling methods applied to the same text-to-image model. `HiRes.Fix' refers to the use of CodeFormer \cite{zhou2022codeformer} as a super-resolution model, employing the method mentioned in \cref{sec:superresolution}. `MultiDiffusion \cite{bar2023multidiffusion}' was used to generate samples via a panoramic approach. The numbers below each column title represent the average elapsed time to generate one sample on an RTX 3090 GPU. UG demonstrates not only competitive quality and fidelity but also relatively faster generation speeds. }
\end{center}
\vskip -0.2in    
\end{figure}

\newpage
\section{Grid Images for Varying Guidance Scale}
\label{apdx:grid}

\begin{figure}[!hb]
\begin{center}
    \centerline{\includegraphics[width=0.9\columnwidth ]{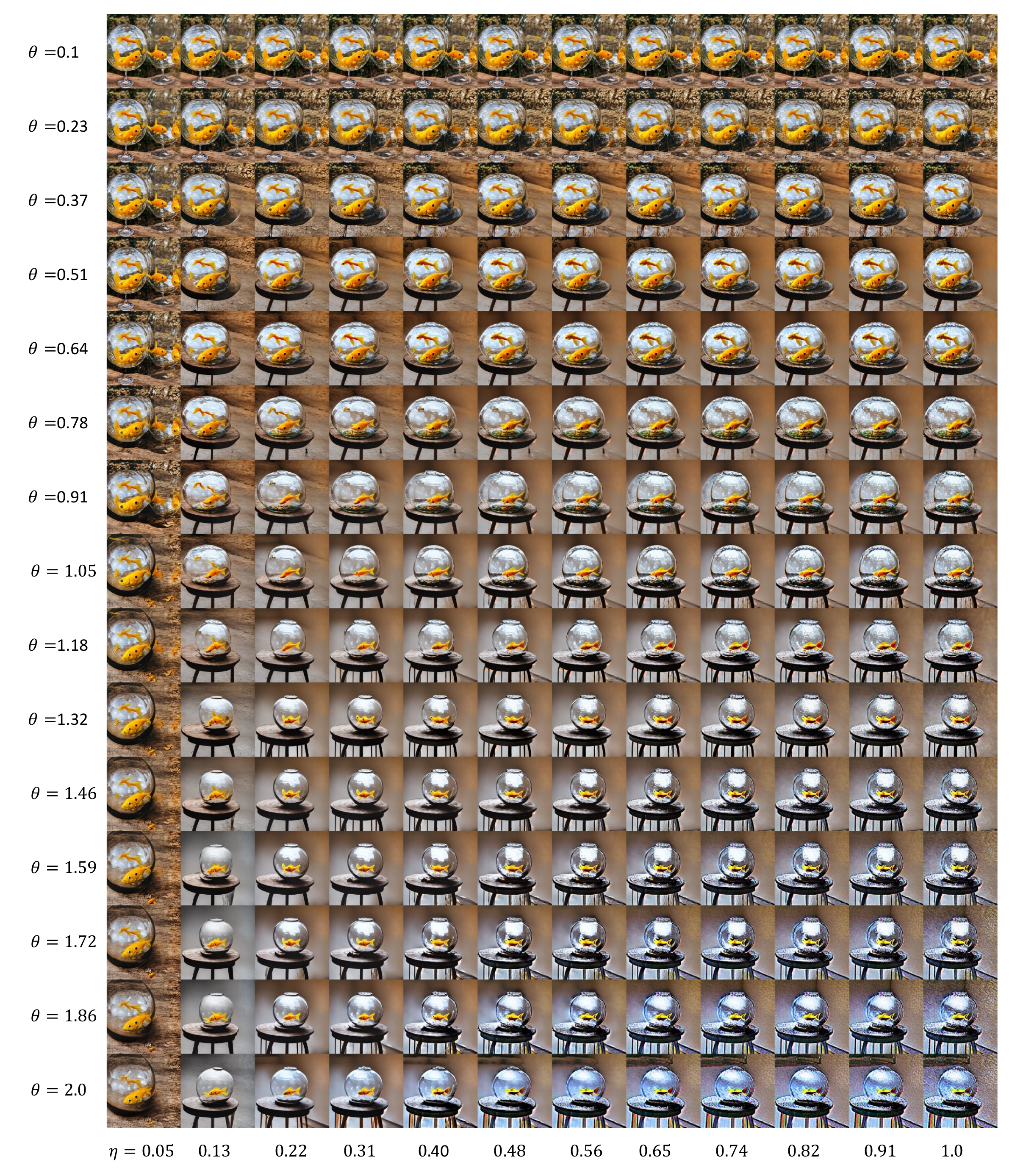}}
    \caption{
Full images obtained by varying the parameters involving the guidance scale in the experiments from \cref{sec:ldm}}
\end{center}
\vskip -0.2in    
\end{figure}

\end{document}